\documentclass{article}

\usepackage{microtype}
\usepackage{graphicx}
\usepackage{subcaption}
\usepackage{booktabs} 
\usepackage{multirow} 

\usepackage{hyperref}
\usepackage{bbm}

\usepackage{amd}

\usepackage{amsmath}
\usepackage{amssymb}
\usepackage{mathtools}
\usepackage{amsthm}

\usepackage{graphicx}
\usepackage{booktabs}
\usepackage{caption}
\usepackage[table]{xcolor}
\usepackage{tcolorbox}
\usepackage{tabularx} 
\usepackage{array}
\usepackage[labelfont=bf,justification=centering]{caption}

\usepackage{microtype}
\usepackage{hyperref}
\usepackage{url}
\usepackage{booktabs}
\usepackage{multirow}
\usepackage[table]{xcolor}
\usepackage{arydshln}
\usepackage{tabularx}
 \usepackage{amsmath}
 \usepackage{tabularx,colortbl}
\usepackage{graphicx}
 \usepackage{xspace}
\usepackage{times}
\usepackage{latexsym}
\usepackage{soul}
\usepackage{tabularx,colortbl}
\usepackage{tabularx}
\usepackage{makecell}
\usepackage{multirow}
\usepackage{float}
\usepackage{arydshln} 
\usepackage{wrapfig}
\usepackage{graphics}
\usepackage{graphicx}
\usepackage{listings}
\usepackage{amssymb}
\usepackage{enumitem}
\usepackage{tcolorbox}
\usepackage{subcaption}
\usepackage{makecell}

\usepackage{fontawesome5}

\usepackage[capitalize,noabbrev]{cleveref}
\usepackage{tikz}
\usepackage{xspace}
\usepackage{xspace}

\newcommand{\AMDlogo}{\includegraphics[height=0.28in]{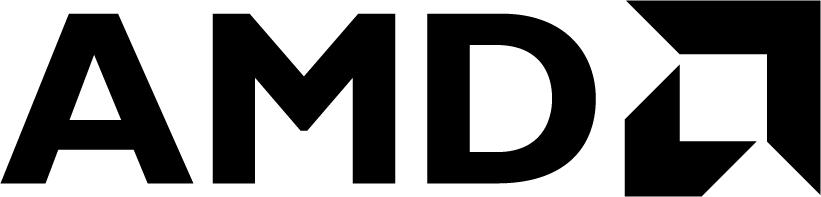}}

\theoremstyle{plain}

\theoremstyle{definition}

\theoremstyle{remark}

\usepackage[textsize=tiny]{todonotes}

\usepackage{lineno}

\definecolor{darkblue}{rgb}{0, 0, 0.5}
\hypersetup{colorlinks=true, citecolor=darkblue, linkcolor=darkblue, urlcolor=darkblue}
\definecolor{oursgray}{RGB}{245,245,245}
\definecolor{oursgray}{gray}{0.94}

\amdtitlerunning{AgentKernelArena: Generalization-Aware Benchmarking of GPU Kernel Optimization Agents}

\begin{document}

\onecolumn

  \amdtitle{AgentKernelArena: Generalization-Aware Benchmarking of GPU Kernel Optimization Agents}




  \begin{amdauthorlist}
    \amdauthor{Sharareh Younesian}{equal} 
    \amdauthor{Wenwen Ouyang}{equal}
    \amdauthor{Sina Rafati}{equal}
    \amdauthor{Mehdi Rezagholizadeh}{equal} 
    \amdauthor{Sharon Zhou}{equal}\\
    \amdauthor{Ji Liu}{}
    \amdauthor{Yue Liu}{}
    \amdauthor{Yuchen Yang}{}
    \amdauthor{Hao Li}{}
    \amdauthor{Ziqiong Liu}{}
    \amdauthor{Dong Li}{}
    \amdauthor{Vikram Appia}{}
    \amdauthor{Zhenyu Gu}{}
    \amdauthor{Emad Barsoum}{}
  \end{amdauthorlist}

  \amdaffiliation{}{AMD}

\begin{center}
     AMD
    \vspace{-20pt}
\end{center}

\amdcorrespondingauthor{Correspondence}{\{sharareh.younesian, vincent.ouyang, sina.rafati, mehdi.rezagholizadeh, sharon.zhou\}@amd.com}


  \vskip 0.3in




\amdCoreContribution

\begin{abstract}
  GPU kernel optimization is increasingly critical for efficient deep learning systems, but writing high-performance kernels still requires substantial low-level expertise. Recent AI coding agents can iteratively read code, invoke compilers and profilers, and refine implementations, yet existing kernel benchmarks evaluate single LLM calls rather than full agent workflows, and none include both kernel-to-kernel optimization and unseen-configuration generalization testing. We present \textbf{AgentKernelArena}, an open-source benchmark for measuring AI coding agents on GPU kernel optimization. The benchmark contains 196 tasks spanning HIP-to-HIP optimization, Triton-to-Triton optimization, and PyTorch-to-HIP translation, and evaluates complete agent workflows in isolated workspaces using gated compilation, correctness, and performance checks, centralized scoring and an unseen-configuration generalization protocol that tests whether optimizations transfer to input configurations the agent never observed. Across production agents including Cursor Agent, Claude Code, and Codex Agent, we find near-perfect compilation and high correctness rates on most task categories, with the strongest configurations achieving mean speedups of up to 6.89$\times$ on PyTorch-to-HIP, 6.69$\times$ on HIP-to-HIP, and 2.13$\times$ on Triton-to-Triton tasks. Our unseen-configuration evaluation shows that HIP-to-HIP and Triton-to-Triton optimizations largely transfer to unseen input shapes, while PyTorch-to-HIP exhibits substantial correctness drops, indicating that agents generating kernels from scratch frequently hardcode shape-specific assumptions. AgentKernelArena is designed as a modular, extensible framework for rigorous evaluation of agentic GPU kernel optimization across agents, tasks, and hardware targets.  
  \\  
  \\
  \noindent\href{https://github.com/AMD-AGI/AgentKernelArena}{\faGithub\ Code: \texttt{github.com/AMD-AGI/AgentKernelArena}}

\end{abstract}

\section{Introduction}
\label{sec:intro}

GPU kernel optimization is central to the performance of modern deep learning systems. As models grow in scale and inference costs dominate deployment budgets, the ability to write fast, correct GPU kernels across programming models and hardware backends has become a critical bottleneck. Traditionally, this work requires deep hardware expertise: understanding memory hierarchies, parallel execution models, instruction selection, and architecture-specific features such as specialized matrix-multiply units, tensor acceleration hardware, and low-level scheduling behavior.

Recent advances in AI coding agents, autonomous systems that can read code, invoke compilers and profilers, and iteratively refine their output, suggest a new approach to kernel optimization.
Rather than relying on a single LLM generation, these agents engage in multi-turn development loops that mirror how human engineers work: write, compile, test, profile, and iterate.
Production tools such as Cursor Agent~\citep{cursoragent}, Claude Code~\citep{claudecode}, and OpenAI Codex~\citep{openaicodex} already support this workflow.
Existing code benchmarks, however, do not measure how well these agents optimize GPU kernels: SWE-bench~\citep{swe-bench} targets general software engineering, HumanEval~\citep{humaneval} scores single-shot code generation, and KernelBench~\citep{kernelbench}, TritonBench~\citep{li2025tritonbench}, and robust-kbench~\citep{robustkbench} evaluate kernel generation from a specification via single LLM calls or light iterative prompting, with no tool-using agent loop and no kernel-to-kernel optimization setting.
None of them test whether agent-produced optimizations generalize to unseen input configurations the agent did not see.

We introduce \textbf{AgentKernelArena}, an open-source evaluation arena for benchmarking AI coding agents on GPU kernel optimization tasks.
Our contributions are:

\begin{enumerate}[leftmargin=*]
  \item \textbf{An agent-centric benchmark} with 196 tasks across three categories (HIP-to-HIP, Triton-to-Triton, PyTorch-to-HIP). Each agent runs in a sandboxed workspace and is evaluated through a  compile $\to$ correctness $\to$ performance gating pipeline, rather than scoring isolated LLM outputs.
  \item \textbf{A centralized evaluation framework} that separates kernel optimization from scoring, enabling fair and reproducible comparison across heterogeneous agent architectures.
  \item \textbf{An unseen-configuration generalization protocol for agentic code generation} that is to our knowledge, the first evaluation that systematically tests whether agent-optimized GPU kernels transfer to unseen input configurations, revealing that agents frequently hardcode shape-specific assumptions that break on inputs they never saw.
  \item \textbf{A modular, extensible design} where new agents, tasks, and hardware targets can be added via configuration, lowering the barrier for the community to benchmark kernel optimization agents.
\end{enumerate}

\section{Related Work}
\label{sec:related}

\paragraph{Code generation and agent benchmarks.}
HumanEval~\citep{humaneval} and MBPP~\citep{mbpp} measure functional correctness of LLM-generated Python on short function-level problems.
SWE-bench~\citep{swe-bench} and AgentBench~\citep{agentbench} extend evaluation to repository-level patches and multi-environment agentic tasks, but target general software engineering rather than performance-critical GPU programming.

\paragraph{GPU kernel benchmarks.}
A growing family of benchmarks evaluates LLM-based kernel generation.
KernelBench~\citep{kernelbench} evaluates kernel generation from PyTorch specifications across 250 tasks and introduces the $\text{fast}_p$ speedup metric;
TritonBench~\citep{li2025tritonbench} targets Triton kernel generation with code-similarity, accuracy, and speedup channels;
ROCmBench~\citep{geak} provides Triton tasks on AMD GPUs;
robust-kbench~\citep{robustkbench} addresses correctness-cheating in prior CUDA benchmarks via LLM-based verifiers and robustness filters;
and MultiKernelBench~\citep{multikernelbench} extends kernel evaluation to multiple hardware platforms.
AgentKernelArena complements this line of work in three ways: (i) it evaluates agents that autonomously compile, test, and profile inside a sandboxed workspace across multiple turns, rather than scoring isolated LLM outputs; (ii) it adds kernel-to-kernel optimization tasks (HIP-to-HIP, Triton-to-Triton) alongside the generation tasks (PyTorch-to-HIP); and (iii) it adds unseen input shapes that test whether reported speedups generalize beyond the configurations the agent saw during optimization.

\paragraph{LLM-driven kernel optimization systems.}
Several recent systems use LLMs to optimize or generate GPU kernels, forming the class of methods that benchmarks like ours are designed to evaluate: QiMeng-Kernel~\citep{qimengkernel}, AutoTriton~\citep{autotriton}, TritonForge~\citep{tritonforge}, AdaExplore~\citep{adaexplore}, and GEAK~\citep{geak}.
Each system is reported under a different evaluation protocol, making cross-system comparison difficult; AgentKernelArena provides a standardized arena into which such systems can be plugged as new agent entries.

\paragraph{AI coding agents.}
Coding agents (SWE-agent~\citep{yang2024swe}, Cursor Agent, Claude Code, OpenAI Codex) have shifted the focus from single-shot generation to multi-turn, tool-augmented development, with substantially higher success rates on complex tasks~\citep{swe-bench}.
AgentKernelArena provides a domain-specific benchmark for these agents on GPU kernel optimization, where iterative compilation and profiling feedback is particularly valuable.

\section{AgentKernelArena: An Arena for Evaluating GPU Kernel Optimization Agents}
\label{sec:framework}

AgentKernelArena is an open-source evaluation arena for measuring how well AI coding agents perform on GPU kernel optimization tasks.
Unlike prior work that evaluates single-shot or iterative LLM calls~\citep{kernelbench}, AgentKernelArena evaluates full agentic systems in a siloed benchmarking environment where each agent is given a real kernel optimization problem, a complete development workspace, and the freedom to compile, test, profile, and iterate autonomously. Moreover, to our knowledge, AgentKernelArena is the first benchmark to systematically evaluate the unseen-configuration generalization of agent-generated GPU kernels, exposing whether reported correctness and speedups survive on input configurations the agent never saw or merely reflect overfitting to visible test configurations.

\subsection{Benchmark Design}
\label{sec:design}

\paragraph{Agent-centric evaluation.}
AgentKernelArena evaluates agents that iteratively modify kernel code.
Each agent receives the same prompt comprising the task type, source files to modify, target kernel functions, compile/correctness/performance commands, optional cheatsheets, and workspace path.
The agent operates in the workspace with full shell access and may iterate autonomously for up to a configurable timeout.
The prompt further instructs the agent to produce up to \texttt{max\_iterations} successive versions of the kernel; this is delivered as a natural-language directive appended to the prompt, rather than a hard runtime cap on tool calls. Agents are free to internally perform more tool invocations between versions.

\paragraph{Domain-specific cheatsheets.}
Optionally, agents receive hardware-specific reference material: a GPU architecture guide,
a HIP best practices document, and a Triton best practices document.
Cheatsheets are user-configurable per task type and per GPU architecture, and are appended verbatim to the agent prompt when enabled (\S\ref{app:prompt}).

\paragraph{Workspace isolation.}
Each task execution creates a timestamped, isolated workspace containing a complete copy of the task source files, evaluation scripts, and build infrastructure; agents cannot access other tasks, prior runs, or other agents' results.
This ensures reproducibility, prevents shared-state corruption, and enables parallel multi-GPU evaluation.

\begin{figure}[t]
  \centering
  \includegraphics[width=0.80\linewidth]{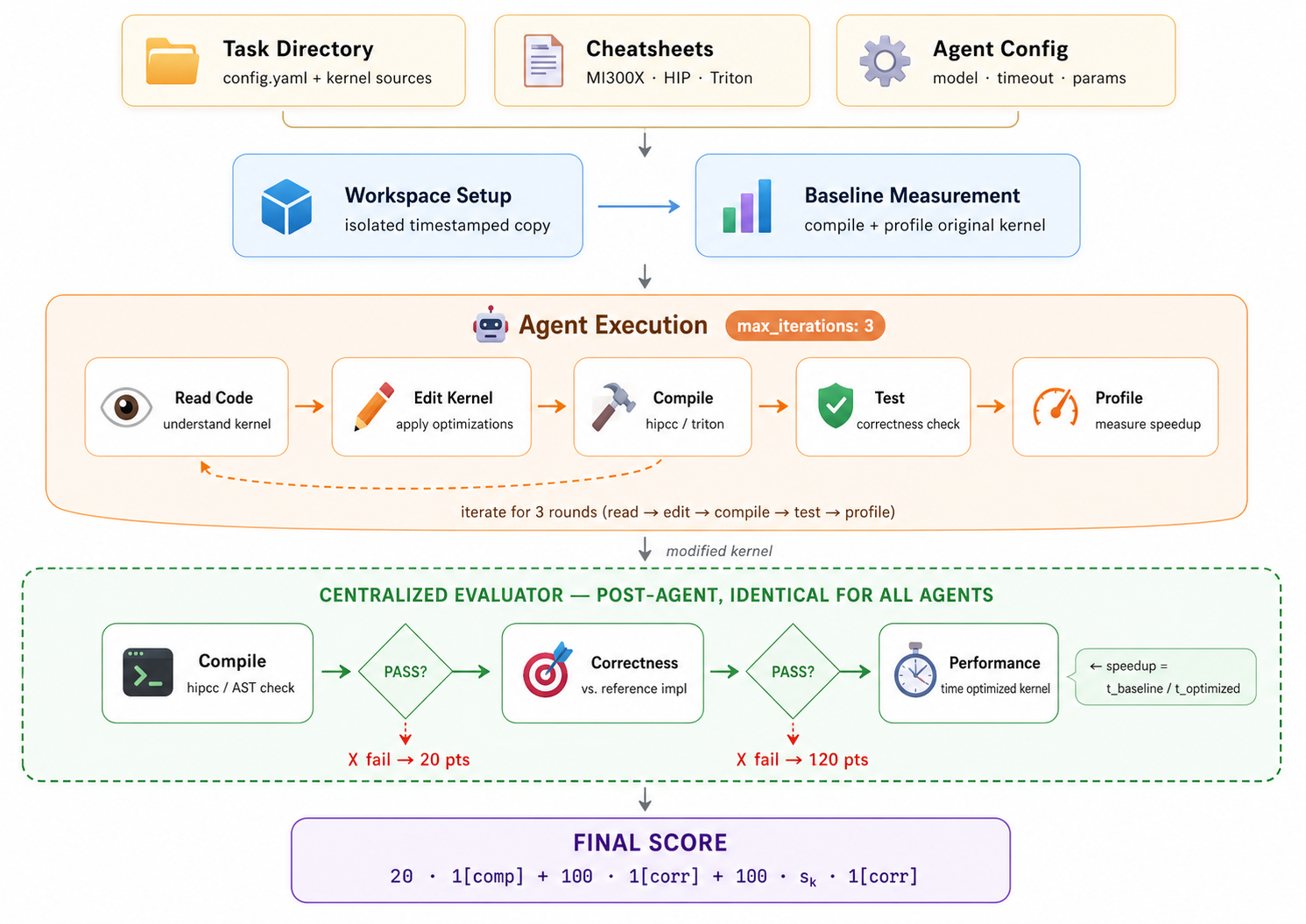}
  \caption{AgentKernelArena evaluation pipeline. \textbf{Top:} task source files, optional cheatsheets, and agent configuration are inputs.
  \textbf{Middle:} the workspace is set up, the original kernel is baselined, and the agent iteratively optimizes the kernel -- prompted to produce up to \texttt{max\_iterations} successive versions (default 3).
  \textbf{Bottom:} after the agent session ends, a centralized evaluator independently runs gated compilation, correctness (vs.\ reference), and performance measurement on the optimized kernel. Speedup is computed as $t_{\text{baseline}} / t_{\text{optimized}}$.
  The scoring function (Eq.~\ref{eq:score}) assigns 20 points for compilation, 100 for correctness, and $100 \cdot s_k$ for performance.}
  \label{fig:framework}
\end{figure}

\paragraph{Execution flow.}
For each task the pipeline proceeds as: (1) workspace setup, isolating the task in a timestamped directory; (2) baseline measurement, compiling the original kernel and profiling its performance; (3) agent execution, launching the agent with a configurable timeout; (4) centralized evaluation, compiling, testing, and profiling the agent's modified kernel with the same commands used for the baseline.
Evaluation is strictly gated: correctness runs only if compilation succeeds, and performance profiling runs only if correctness passes.
 Speedup is computed by arithmetic averaging of per test-case speedup ratios.
Figure~\ref{fig:framework} illustrates this pipeline.

\subsection{Task Selection}
\label{sec:tasks}

AgentKernelArena comprises 196 tasks drawn from real-world GPU workloads, organized into three core categories by task type.
Tasks are sourced from production ML codebases and open-source GPU kernel repositories, ensuring that progress on the benchmark translates to practical impact.
Table~\ref{tab:tasks} summarizes the task categories.

\paragraph{Task categories.}
We define three task types based on the source and target programming models:

\begin{itemize}
  \item \textbf{HIP-to-HIP} (24 tasks). The agent receives a reference HIP kernel and must produce an optimized version.
  Tasks are drawn from the GPU Mode community~\citep{gpumode} and cover activations (GELU, SiLU, Sigmoid), attention mechanisms (multi-head, dot-product), normalization layers (LayerNorm, BatchNorm), matrix operations, and loss functions.
  Correctness is evaluated by comparing PyTorch module output against a functional path that injects the agent's compiled HIP kernel; performance is measured as speedup over a provided reference HIP implementation.
  These tasks test the agent's ability to apply GPU-specific optimizations
  to existing kernel code.

  \item \textbf{Triton-to-Triton} (148 tasks). The agent receives a reference Triton kernel and must produce a faster version.
  This category draws from two sources: 118 kernels from the vLLM inference engine~\citep{vllm} (attention, mixture-of-experts routing, quantization, memory management, sampling) and 30 kernels from ROCmBench~\citep{geak} covering element-wise operations, reductions, normalization, GEMM variants, flash attention, and MoE kernels.
  Triton's block-level programming model shifts the optimization space toward block size tuning, fusion strategies, and memory access pattern optimization.

  \item \textbf{PyTorch-to-HIP} (24 tasks). The agent receives a PyTorch \texttt{nn.Module} as specification and must create an equivalent HIP kernel from scratch; no reference HIP file is provided.
  This is the most demanding category: the agent must bridge the abstraction gap between a high-level functional specification and low-level GPU code, handling memory layout, thread mapping, and numerical precision.
  Correctness is verified against the PyTorch module output, and performance is measured as the speedup of the agent's HIP kernel over PyTorch eager execution.
  Tasks mirror the HIP-to-HIP operator set (GELU, SiLU, softmax, multi-head attention, etc.).
\end{itemize}

\begin{table}[t]
  \caption{Task categories in AgentKernelArena. Each task is self-contained with its own compilation, correctness, and performance evaluation scripts, validated by an automated task validator agent.}
  \label{tab:tasks}
  \centering\small
  \resizebox{0.85\textwidth}{!}{%
  \begin{tabular}{llrl}
    \toprule
    Category & Source & Tasks & Example kernels \\
    \midrule
    HIP-to-HIP & GPU Mode community & 24 & GELU, MultiHeadAttention, LayerNorm \\
    Triton-to-Triton (vLLM) & vLLM inference & 118 & fused MoE, scaled MM, paged decode \\
    Triton-to-Triton (ROCmBench) & ROCmBench & 30 & flash attention, GEMM, softmax \\
    PyTorch-to-HIP & GPU Mode community & 24 & SiLU, Softmax, Transformer FFN \\
    \midrule
    \textbf{Total} & & \textbf{196} & \\
    \bottomrule
  \end{tabular}%
  }
\end{table}

\paragraph{Multi-shape evaluation.}
Unlike benchmarks that evaluate on a single fixed input shape (e.g. in ~\citep{kernelbench}), each task includes multiple input configurations that are visible to the agent during optimization.
Exposing diverse shapes during optimization encourages agents to produce kernels that are robust across input geometries rather than tuned to a single size. This is distinct from the unseen-configuration generalization protocol below, which evaluates on configurations the agent never sees.

\paragraph{Unseen-configuration generalization evaluation.}
To test whether agents actually generalize or simply hardcode optimizations for the visible shapes, we introduce an unseen-configuration generalization protocol.
For each task, we generate a set of distinct unseen input configurations (e.g., non-power-of-two dimensions or higher-rank tensors) that are never shown to the agent.
After optimization, the kernel is evaluated on both the original and unseen configurations, and we report the generalization gap: $\Delta_g = (\bar{s}_{\text{seen}} - \bar{s}_{\text{unseen}}) / \bar{s}_{\text{seen}}$, where $\bar{s}$ denotes mean speedup.
A small $\Delta_g$ indicates genuine optimization strategies; a large gap suggests overfitting to the visible test shapes.

\subsection{Metrics}
\label{sec:metrics}

We evaluate agent-generated kernels along three axes (compilation, correctness, and performance) and combine them into a unified scoring system that rewards both reliability and optimization quality.

\paragraph{Three-phase evaluation.}
Each submitted kernel is evaluated through a gated pipeline:
\begin{enumerate}
  \item \textbf{Compilation.} The kernel must compile without errors via the task-specific toolchain (\texttt{hipcc} for HIP, AST validation and import for Triton).
  \item \textbf{Correctness.} The compiled kernel must produce outputs matching a reference implementation across all input shapes. References are task-specific: PyTorch module output (HIP-to-HIP, PyTorch-to-HIP) or explicit reference functions (Triton-to-Triton). Tolerances vary by data type and task.
  \item \textbf{Performance.} Execution time is measured with 10 warmup and 100 timed iterations using \texttt{torch.cuda.Event}-based GPU timing. Speedup is $s = t_{\text{base}} / t_{\text{opt}}$, where the baseline is a reference HIP kernel (HIP-to-HIP), PyTorch eager execution (PyTorch-to-HIP), or the unmodified Triton kernel (Triton-to-Triton).
\end{enumerate}

\paragraph{Scoring.}
We use a cumulative scoring function that assigns credit at each evaluation gate:
\begin{equation}
  \text{Score}(k) =
    \underbrace{20 \cdot \mathbbm{1}[\text{compiles}]}_{\text{compilation}} +
    \underbrace{100 \cdot \mathbbm{1}[\text{correct}]}_{\text{correctness}} +
    \underbrace{100 \cdot s_k \cdot \mathbbm{1}[\text{correct}]}_{\text{performance}}
  \label{eq:score}
\end{equation}
where $s_k$ is the speedup ratio for kernel $k$.
Concretely, a compile-only kernel scores 20 points, a correct kernel that merely matches baseline ($s_k{=}1$) scores 220, and a $2\times$ kernel scores 420.
The weights are chosen so that (i)~compilation credit cannot offset a correctness failure, (ii)~any correct kernel strictly dominates any incorrect submission regardless of putative speedup, and (iii)~the linear performance term distinguishes speedups without saturating, unlike bounded metrics such as $\text{fast}_p$.
For multi-shape tasks, $s_k$ is the arithmetic mean of per-shape speedup ratios.

\paragraph{Aggregate metrics.}
To compare agents across the full benchmark, we report:
\begin{itemize}
  \item \textbf{Compilation rate}: fraction of tasks where the agent's kernel compiles.
  \item \textbf{Correctness rate}: fraction of tasks where the kernel passes all correctness checks.
  \item \textbf{Mean speedup} $\pm\,\sigma_r$: arithmetic mean of per-task speedup ratios across all tasks (including 0.0$\times$ for tasks that fail compilation or correctness), with run-to-run standard deviation.
  \item \textbf{Mean score}: arithmetic mean of $\text{Score}(k)$ across all tasks.
  \item \textbf{Geometric mean} $\pm\,\sigma_r$: geometric mean of per-task speedup ratios, computed over correct tasks only (speedup $> 0$). Less sensitive to outlier speedups than the arithmetic mean.
  \item $\textbf{fast}_p$ \textbf{(\%)}: fraction of all tasks achieving speedup $\geq p\times$. We report $p \in \{1, 2\}$ for comparability with KernelBench~\citep{kernelbench}.
  \item \textbf{Unseen-input generalization gap} ($\Delta_g$): mean speedup loss when moving from seen input configurations to unseen ones (see \S\ref{sec:tasks}).
\end{itemize}
Results are reported across the per task category, since evaluation methodologies differ across categories.
We run each agent three times per task and report mean $\pm\,\sigma_r$ to account for non-determinism in both agent behavior and GPU timing; $\sigma_r$ is computed over the 3 runs' aggregate mean speedups and captures run-to-run variability.
This should not be confused with the cross-task speedup distribution reported in Appendix~\ref{app:speedup-dist}, which captures variance across tasks within a single aggregate.

\section{Experiments and Results}
\label{sec:experiments}

We evaluate three production agents (Cursor Agent, Claude Code, and Codex Agent) each with multiple underlying models, across all 196 tasks.
Every configuration is run three times; we average each task's metrics across runs before computing aggregate statistics.
All experiments run on AMD Instinct MI300X with ROCm~7.1.1, PyTorch~2.10.0, and Triton~3.6.0, using a 3600\,s timeout and \texttt{max\_iterations=3}.
Table~\ref{tab:agent-configs} in the appendix lists the full agent configurations; the human-friendly model names used in the result tables below map to concrete API identifier strings and evaluation windows in Table~\ref{tab:model-versions}.

\subsection{Main Results}

Tables~\ref{tab:res-hip}, \ref{tab:res-triton}, and~\ref{tab:res-torch} report per-category results.

\begin{table}[t]
  \caption{HIP-to-HIP results (24 tasks, 3 runs per configuration, per-task averaged).}
  \label{tab:res-hip}
  \centering\small
  \resizebox{0.9\textwidth}{!}{%
  \begin{tabular}{llrr rrr rr}
    \toprule
    Agent & Model & Comp.\,\% & Corr.\,\%
      & Mean Spd.\,$\pm\,\sigma_r$ & Mean Score & Geo.\,Mean\,$\pm\,\sigma_r$
      & $\text{fast}_1$\,\% & $\text{fast}_2$\,\% \\
    \midrule
    Claude Code  & Opus 4.6            & 100.0 &  98.6 & 6.69$\pm$0.51$\times$ &  787.8 & 3.31$\pm$0.13$\times$ & 100.0 & 50.0 \\
    Claude Code  & Sonnet 4.6          & 100.0 & 100.0 & 5.37$\pm$0.43$\times$ &  656.9 & 2.81$\pm$0.08$\times$ &  87.5 & 50.0 \\
    \midrule
    Cursor Agent & Opus 4.7 High       & 100.0 &  95.8 & 5.82$\pm$0.64$\times$ &  698.3 & 2.97$\pm$0.10$\times$ &  95.8 & 45.8 \\
    Cursor Agent & Opus 4.6 High       & 100.0 & 100.0 & 4.62$\pm$0.48$\times$ &  582.3 & 2.42$\pm$0.08$\times$ &  91.7 & 45.8 \\
    Cursor Agent & GPT-5.4 High        & 100.0 & 100.0 & 5.19$\pm$0.36$\times$ &  639.3 & 2.67$\pm$0.21$\times$ &  87.5 & 41.7 \\
    Cursor Agent & GPT-5.3-Codex High  & 100.0 & 100.0 & 3.65$\pm$0.94$\times$ &  484.9 & 2.15$\pm$0.15$\times$ &  87.5 & 41.7 \\
    Cursor Agent & Composer 2          & 100.0 &  98.6 & 1.44$\pm$0.37$\times$ &  262.4 & 1.33$\pm$0.18$\times$ &  83.3 & 12.5 \\
    \midrule
    Codex Agent  & GPT-5.3-Codex       & 100.0 & 100.0 & 3.61$\pm$1.05$\times$ &  480.9 & 2.28$\pm$0.29$\times$ &  95.8 & 45.8 \\
    \bottomrule
  \end{tabular}%
  }
\end{table}

\begin{table}[t]
  \caption{Triton-to-Triton results (148 tasks, 3 runs per configuration, per-task averaged).}
  \label{tab:res-triton}
  \centering\small
  \resizebox{0.9\textwidth}{!}{%
  \begin{tabular}{llrr rrr rr}
    \toprule
    Agent & Model & Comp.\,\% & Corr.\,\%
      & Mean Spd.\,$\pm\,\sigma_r$ & Mean Score & Geo.\,Mean\,$\pm\,\sigma_r$
      & $\text{fast}_1$\,\% & $\text{fast}_2$\,\% \\
    \midrule
    Claude Code  & Opus 4.6            & 100.0 &  99.8 & 2.11$\pm$0.01$\times$ & 331.2 & 1.30$\pm$0.01$\times$ & 86.5 &  9.5 \\
    Claude Code  & Sonnet 4.6          & 100.0 & 100.0 & 2.00$\pm$0.72$\times$ & 320.3 & 1.26$\pm$0.02$\times$ & 77.0 &  8.1 \\
    \midrule
    Cursor Agent & Opus 4.7 High       & 100.0 & 100.0 & 2.13$\pm$0.10$\times$ & 333.2 & 1.31$\pm$0.02$\times$ & 84.5 & 10.1 \\
    Cursor Agent & Opus 4.6 High       & 100.0 & 100.0 & 1.73$\pm$0.19$\times$ & 293.2 & 1.18$\pm$0.01$\times$ & 70.9 &  4.7 \\
    Cursor Agent & GPT-5.4 High        & 100.0 & 100.0 & 1.75$\pm$0.13$\times$ & 295.3 & 1.10$\pm$0.02$\times$ & 52.7 &  3.4 \\
    Cursor Agent & GPT-5.3-Codex High  & 100.0 &  99.3 & 1.65$\pm$0.10$\times$ & 283.8 & 1.04$\pm$0.01$\times$ & 52.0 &  1.4 \\
    Cursor Agent & Composer 2          & 100.0 &  97.7 & 1.59$\pm$0.33$\times$ & 276.8 & 1.01$\pm$0.02$\times$ & 40.5 &  4.1 \\
    \midrule
    Codex Agent  & GPT-5.3-Codex       & 100.0 &  99.3 & 1.68$\pm$0.05$\times$ & 287.5 & 1.06$\pm$0.01$\times$ & 56.1 &  2.0 \\
    \bottomrule
  \end{tabular}%
  }
\end{table}

\begin{table}[t]
  \caption{PyTorch-to-HIP results (24 tasks, 3 runs per configuration, per-task averaged).}
  \label{tab:res-torch}
  \centering\small
  \resizebox{0.9\textwidth}{!}{%
  \begin{tabular}{llrr rrr rr}
    \toprule
    Agent & Model & Comp.\,\% & Corr.\,\%
      & Mean Spd.\,$\pm\,\sigma_r$ & Mean Score & Geo.\,Mean\,$\pm\,\sigma_r$
      & $\text{fast}_1$\,\% & $\text{fast}_2$\,\% \\
    \midrule
    Claude Code  & Opus 4.6            &  98.6 &  97.2 & 6.70$\pm$0.17$\times$ & 787.1 & 4.53$\pm$0.11$\times$ & 100.0 & 79.2 \\
    Claude Code  & Sonnet 4.6          & 100.0 & 100.0 & 5.30$\pm$0.35$\times$ & 649.6 & 3.79$\pm$0.20$\times$ & 100.0 & 83.3 \\
    \midrule
    Cursor Agent & Opus 4.7 High       & 100.0 & 100.0 & 6.65$\pm$0.44$\times$ & 785.1 & 4.64$\pm$0.14$\times$ & 100.0 & 83.3 \\
    Cursor Agent & Opus 4.6 High       & 100.0 &  98.6 & 6.89$\pm$1.15$\times$ & 807.2 & 4.49$\pm$0.47$\times$ & 100.0 & 75.0 \\
    Cursor Agent & GPT-5.4 High        &  69.4 &  69.4 & 3.85$\pm$0.59$\times$ & 468.3 & 2.19$\pm$0.88$\times$ &  75.0 & 58.3 \\
    Cursor Agent & GPT-5.3-Codex High  &  93.1 &  88.9 & 3.74$\pm$0.40$\times$ & 481.7 & 2.76$\pm$0.15$\times$ &  87.5 & 62.5 \\
    Cursor Agent & Composer 2          & 100.0 & 100.0 & 4.14$\pm$0.68$\times$ & 534.3 & 3.05$\pm$0.57$\times$ &  95.8 & 70.8 \\
    \midrule
    Codex Agent  & GPT-5.3-Codex       & 100.0 & 100.0 & 5.20$\pm$0.22$\times$ & 640.2 & 3.79$\pm$0.26$\times$ & 100.0 & 75.0 \\
    \bottomrule
  \end{tabular}%
  }
\end{table}

\paragraph{Compilation and correctness.}
All configurations achieve near-perfect compilation rates across all categories.
The one notable exception is Cursor Agent with GPT-5.4 High on PyTorch-to-HIP, where compilation drops to 69.4\%, the lowest compilation rate observed across all configurations for this category.
Correctness rates are uniformly high for HIP-to-HIP and Triton-to-Triton (${\geq}91\%$), indicating that agents reliably preserve functional equivalence when optimizing existing kernels.

\begin{figure}[t]
  \centering
  \includegraphics[width=0.65\linewidth]{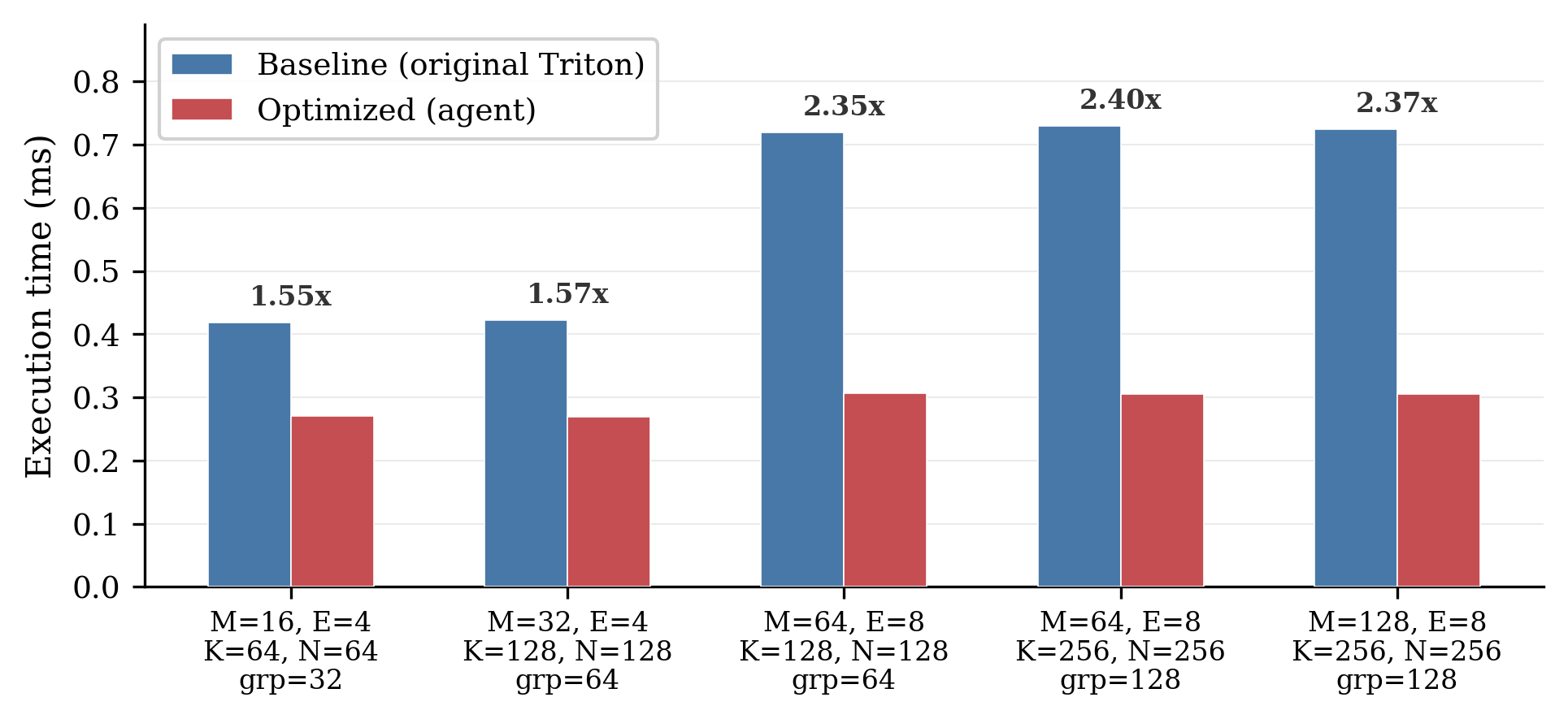}
  \caption{Per-test-case execution time comparison for the \texttt{fused\_moe\_gptq\_awq} kernel (Triton-to-Triton, Claude Code / Opus~4.6). Each bar pair shows baseline vs.\ optimized execution time for a different parameter configuration (M=tokens, E=experts, K/N=matrix dimensions, grp=quantization group size). The agent achieves 1.55--2.40$\times$ speedup, with larger gains at higher expert counts and matrix dimensions.}
  \label{fig:per-testcase}
\end{figure}

\paragraph{Performance across categories.}
PyTorch-to-HIP yields the highest speedups (mean 3.74--6.89$\times$, geometric mean 2.19--4.64$\times$), since agents generate HIP kernels that replace PyTorch eager execution, a comparatively slow baseline; the top configurations achieve $\text{fast}_2 \geq 75\%$.
HIP-to-HIP shows moderate gains (mean 1.44--6.69$\times$, geometric mean 1.33--3.31$\times$) with high variance, as some kernels (e.g., attention operators) offer significant optimization headroom while others are already well-tuned.
Triton-to-Triton is the most challenging category: mean speedups range from 1.59--2.13$\times$ and geometric means from 1.01--1.31$\times$, reflecting Triton's compiler-managed optimization that leaves less room for manual improvement; $\text{fast}_2$ rates are below 11\% for all configurations.
Figure~\ref{fig:per-testcase} illustrates a representative per-test-case breakdown for a Triton-to-Triton task.

\paragraph{Agent and model rankings.}
Claude Code with Opus 4.6 achieves the highest mean speedup on HIP-to-HIP (6.69$\times$) and is competitive on Triton-to-Triton (2.11$\times$) and PyTorch-to-HIP (6.70$\times$).
Cursor Agent with Opus 4.7 High is the strongest Cursor configuration, ranking first on Triton-to-Triton (2.13$\times$) and achieving the highest geometric mean on PyTorch-to-HIP (4.64$\times$).
Codex Agent with GPT-5.3-Codex performs competitively: 3.61$\times$ on HIP-to-HIP, 5.20$\times$ on PyTorch-to-HIP, and 1.68$\times$ on Triton-to-Triton, comparable to Cursor Agent with similar models.
Within the Cursor Agent Opus 4.7 High and Opus 4.6 High lead across categories, followed by GPT-5.4 High and GPT-5.3-Codex High, with the top two models trading places on PyTorch-to-HIP (Opus 4.6 High achieves 6.89$\times$ vs.\ 6.65$\times$ for Opus 4.7 High).


\subsection{Unseen-Configuration Generalization Analysis}
\label{sec:generalization}

To evaluate whether agent-optimized kernels generalize beyond the input configurations visible during development, we run the unseen-configuration generalization protocol described in \S\ref{sec:tasks} on every  configuration.

\paragraph{Unseen configuration generation.}
For each of the 196 tasks, we use Cursor Agent with \texttt{claude-opus-4-6-high} to inspect the kernel source and existing test infrastructure, then generate 8 structurally diverse unseen configurations spanning six generalization categories:
edge-case/boundary (e.g., batch$=$1, dimension equal to BLOCK\_SIZE), scale-up (${\geq}2{-}4{\times}$ the dominant dimension),
scale-down (${\leq}2{-}4{\times}$),
alignment-stress (prime or non-power-of-two sizes such as 37, 131, 4003),
asymmetric aspect ratio (e.g., $M{=}1, N{=}65536$),
and production-realistic (shapes drawn from real transformer workloads).
Each configuration is tagged with its category, enabling per-category analysis of failure modes.
To prevent contamination of future evaluations, we do not release the unseen configurations; we do release the generation script so that the protocol is fully reproducible and extensible.

\paragraph{Evaluation protocol.}
For each run, the evaluation script injects the same unseen configurations into two workspace copies (one with the agent's optimized kernel, one with the original) and runs both through the standard compile/correctness/performance pipeline.

\paragraph{Generalization quadrant.}
Each task is classified into one of four outcomes: \texttt{both\_pass}, \texttt{opt\_regression} (optimization broke generalization), \texttt{both\_fail} (configuration exceeds kernel design spec), or \texttt{opt\_improvement} (agent improved robustness).
The key metric is conditional correctness: $P(\text{opt correct} \mid \text{orig correct})$, which excludes configurations inherently beyond the kernel's capability.
For \texttt{both\_pass} tasks, we also compute the generalization gap $\Delta_g = (\bar{s}_{\text{seen}} - \bar{s}_{\text{unseen}}) / \bar{s}_{\text{seen}}$.

\begin{figure}[t]
  \centering
  \includegraphics[width=\linewidth]{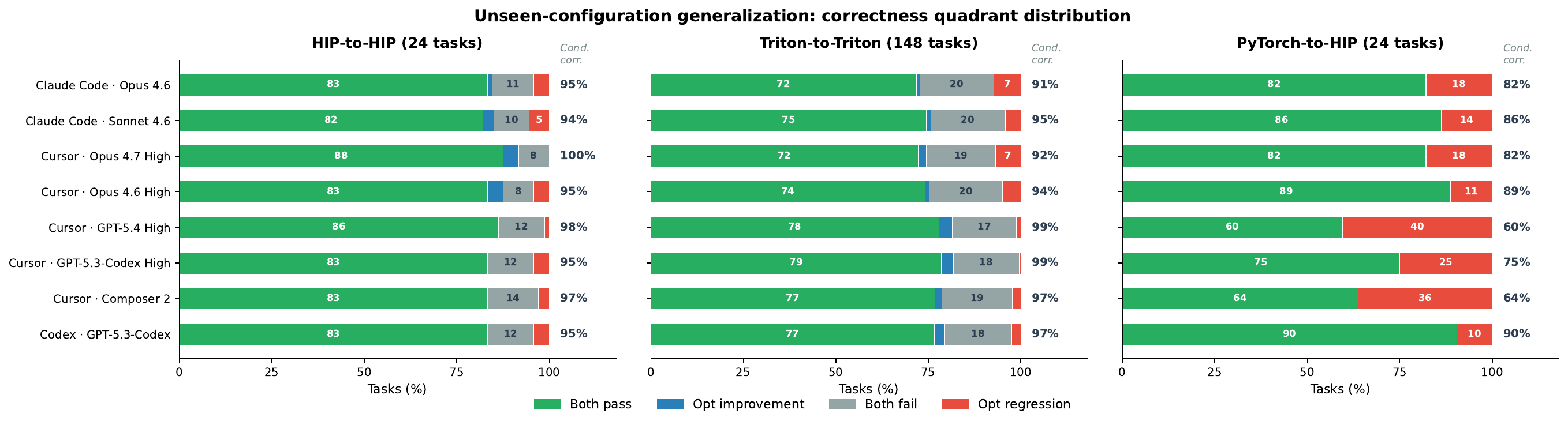}
  \caption{Unseen-configuration generalization: quadrant breakdown. Each horizontal bar shows the fraction of tasks in each correctness quadrant (\texttt{both\_pass}, \texttt{opt\_improvement}, \texttt{both\_fail}, \texttt{opt\_regression}). Conditional correctness (\%) is annotated on the right.}
  \label{fig:heldout-quadrant}
\end{figure}

\begin{figure}[t]
  \centering
  \includegraphics[width=\linewidth]{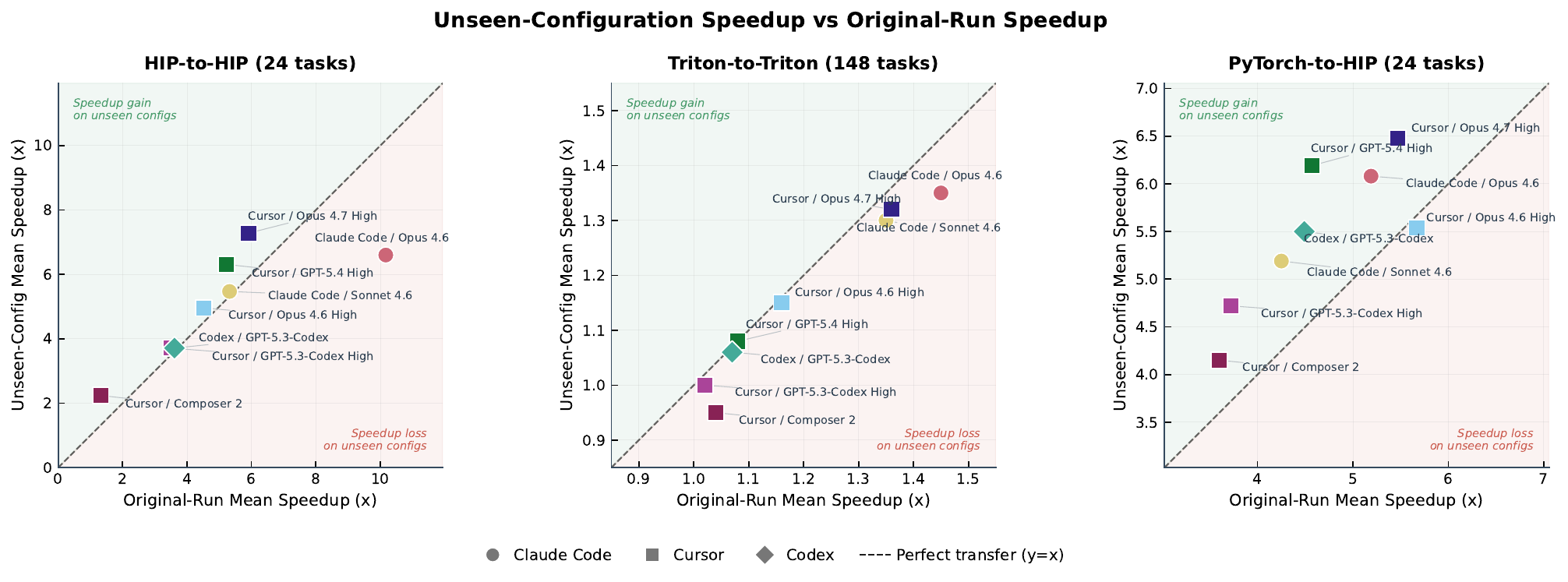}
  \caption{Unseen vs.\ original-run mean speedup, per agent/model and per task category. Marker color encodes the model and marker shape encodes the agent platform. The dashed line is \(y = x\) (perfect transfer); points in the \textcolor[HTML]{1e8449}{green} region above the diagonal generalize better on unseen configurations than on original ones, while points in the \textcolor[HTML]{c0392b}{red} region below the diagonal lose speedup on unseen inputs. The vertical distance from each point to the diagonal equals the absolute generalization gap.}
  \label{fig:heldout-speedup}
\end{figure}

Figures~\ref{fig:heldout-quadrant} and~\ref{fig:heldout-speedup} summarize the unseen-configuration generalization results across all agents and task categories (3 runs averaged per configuration).

\paragraph{HIP-to-HIP generalization.}
As shown in the left panel of Figure~\ref{fig:heldout-quadrant}, HIP kernels generalize well: conditional correctness ranges from 93.6\% to 100\%.
In Figure~\ref{fig:heldout-speedup} (left), most points lie above the diagonal: several configurations gain speedup on unseen inputs (e.g., Cursor / Opus 4.7 High, $+23\%$; Cursor / GPT-5.4 High, $+21\%$), because unseen configurations occasionally expose latent parallelism that the kernel already handles correctly.
Cursor / Opus 4.7 High achieves 100\% conditional correctness with zero regressions, indicating fully shape-agnostic optimizations.

\paragraph{Triton-to-Triton generalization.}
Triton kernels also demonstrate strong generalization (center panels), with conditional correctness between 90.9\% and 99.4\%.
The generalization gaps are small ($|\Delta_g| < 0.1$ for every configuration), confirming that Triton's block-structured programming model naturally constrains optimizations to be shape-general.
Cursor / GPT-5.3-Codex High achieves the highest conditional correctness at 99.4\%, while the \texttt{opt\_improvement} counts indicate that agents sometimes produce kernels that are more robust to novel shapes than the original implementations.

\paragraph{PyTorch-to-HIP generalization.}
PyTorch-to-HIP exhibits the lowest conditional correctness (59.7\% to 90.3\%), as expected: agents generating HIP kernels from scratch are more likely to hardcode dimension-specific assumptions that break on unseen shapes.
Despite lower correctness retention, correctly generalizing kernels run faster on unseen inputs (Figure~\ref{fig:heldout-speedup}, right).
Codex / GPT-5.3-Codex achieves the best balance: 90.3\% conditional correctness with competitive unseen-input speedup (5.50$\times$).
We provide a detailed categorization of failure modes (compilation, correctness, and unseen-input regressions) with representative error examples and per-model susceptibility analysis in Appendix~\ref{app:failures}.

\section{Discussion}
\label{sec:discussion}

\paragraph{Agent behavior patterns.}
All agents follow iterative compile--test--benchmark loops, but their strategies differ by task type.
On \textbf{HIP-to-HIP} tasks, agents apply low-level GPU optimizations: higher-capability models use kernel fusion, vectorized loads (\texttt{float4}), warp-shuffle reductions, and \texttt{\_\_launch\_bounds\_\_} tuning, while lower-capability models default to block-size adjustments and loop unrolling.
On \textbf{Triton-to-Triton} tasks, optimization centers on \texttt{@triton.autotune} configurations, adjusting \texttt{BLOCK\_SIZE}, \texttt{num\_warps}, \texttt{num\_stages}, and AMD-specific knobs like \texttt{waves\_per\_eu}.
On \textbf{PyTorch-to-HIP} tasks, where agents generate HIP kernels from scratch, agents must bridge the abstraction gap from high-level module semantics to thread/block/grid mappings, memory allocation, and Python bindings.
A detailed breakdown is provided in Appendix~\ref{app:behavior}.

\paragraph{Computational cost.}
Claude Code is the most verbose agent, averaging 39--86K output tokens per task; Cursor Agent ranges from 8--25K; and Codex Agent is the most concise at 13--17K (Table~\ref{tab:tokens} in the appendix).
Higher token budgets generally correlate with higher speedups, though the relationship is sublinear.

\paragraph{Scope and limitations.}
The current study targets a single GPU architecture and evaluates three commercial agents with \texttt{max\_iterations=3} and three runs per configuration, both bounded by API cost; sensitivity to larger iteration budgets and to additional hardware is left to future work.
Model availability varies across platforms, so most cross-model comparisons are conducted within the Cursor Agent.
Open-weight models were probed but consistently failed at compilation in single-iteration calls due to the multi-file context required.
Wrapping such models in a comparable iterative agent loop (with shell/compile/profile tool access, error-feedback routing, and retry policies) is a non-trivial engineering effort and is explicitly left to future work.
Specialized kernel optimization systems (e.g., GEAK, AutoTriton) were excluded because their task-specific architectures differ from general-purpose coding agents, although the framework readily supports their integration.

\paragraph{Broader impact.}
By providing a standardized arena for evaluating kernel optimization agents, AgentKernelArena can accelerate the development of AI-assisted GPU programming tools and lower the barrier to high-performance computing.
The framework is designed so that new agents, tasks, and hardware targets can be easily added, requiring only a launcher script and YAML config for agents, a self-contained directory for tasks, and a cheatsheet entry for new GPU architectures (see Appendix~\ref{app:extensibility}), enabling both agent developers and hardware vendors to contribute to and benefit from the benchmark.
Faster AI-assisted kernel optimization could meaningfully lower the cost of training and inference; we view rigorous unseen-configuration generalization evaluation as a prerequisite for responsible adoption of agent-generated kernels in production systems.

\section{Conclusion}

We presented AgentKernelArena, an open-source benchmark for evaluating AI coding agents on GPU kernel optimization.
The benchmark includes 196 tasks spanning HIP-to-HIP optimization, Triton-to-Triton optimization, and PyTorch-to-HIP translation, evaluated through a gated compile--correctness--performance pipeline with centralized scoring.
Our results show that production agents already achieve near-perfect compilation and high correctness, with speedups of up to 6.89$\times$ over baseline; yet our unseen-configuration generalization protocol reveals a critical gap: agents generating kernels from scratch (PyTorch-to-HIP) suffer correctness drops of up to 40\% on unseen input configurations, exposing pervasive shape-specific hardcoding that inflates seen-input metrics.
This finding underscores that standard benchmarks measuring performance only on visible configurations can substantially overestimate the reliability of agent-generated code, and that unseen-input evaluation should be a first-class component of any agentic code generation benchmark.

\bibliography{ref}
\bibliographystyle{ref}

\newpage
\appendix

\section{Task Curation Process}
\label{app:curation}

The 196 tasks in AgentKernelArena were curated from three sources, each requiring different processing pipelines.

\paragraph{HIP-to-HIP and PyTorch-to-HIP tasks (48 tasks).}
The PyTorch module specifications were derived from seed kernels in the GPU Mode community's kernel dataset~\citep{gpumode}, released under the CC-BY-4.0 license.
Starting from these PyTorch modules, we used an LLM-assisted pipeline to generate corresponding HIP kernel implementations, which were then manually reviewed for correctness and performance.
Each task was packaged with evaluation tooling (compilation scripts, correctness checks against PyTorch reference output, and performance measurement scripts) and validated end-to-end on the target hardware.
The same 24 operator set is used for both HIP-to-HIP (where the agent optimizes an existing HIP kernel) and PyTorch-to-HIP (where the agent generates a HIP kernel from scratch).

\paragraph{Triton-to-Triton vLLM tasks (118 tasks).}
Triton kernels were extracted from the vLLM inference engine repository~\citep{vllm}, released under the Apache-2.0 license.
Since vLLM kernels are typically embedded in larger modules with multiple interdependent functions, an LLM agent was used to isolate each kernel into a self-contained single-file format and generate the accompanying scaffold code: a test runner, input generators, correctness checks, and performance measurement scripts.
Each extracted task was validated to compile, pass correctness, and produce meaningful performance baselines on the target hardware.

\paragraph{Triton-to-Triton ROCmBench tasks (30 tasks).}
These tasks were sourced from the ROCmBench evaluation suite~\citep{geak}, which provides Triton kernel tasks targeting AMD GPUs and is released as an open-source repository accompanying the originating paper.
Each task was adapted into the AgentKernelArena task format with the same evaluation tooling as other Triton tasks, with attribution to the originating paper preserved in per-task metadata.

\paragraph{Task validation.}
All 196 tasks were validated using an automated task validator agent that verifies the task directory structure, runs compilation, correctness, and performance scripts, and flags any issues.
Tasks that failed validation were either fixed or excluded.

\section{Complete Task List}
\label{app:tasks}

Table~\ref{tab:full-tasks} lists all 196 tasks in AgentKernelArena, organized by category and source.

\begin{table}[h]
\caption{Complete list of tasks in AgentKernelArena.}
\label{tab:full-tasks}
\centering
\small
\resizebox{0.8\textwidth}{!}
{\begin{tabular}{lll}
\toprule
Category & Task name & Operator type \\
\midrule
\multicolumn{3}{l}{\textbf{HIP-to-HIP} (24 tasks, source: GPU Mode community)} \\
\midrule
& GELU, SiLU, Sigmoid, TanH, FusedLeakyReLU & Activations \\
& MultiHeadAttention, NormalAttention\_dot & Attention \\
& NormalAttention\_embedded\_gaussian, ItemQueryAttention & Attention \\
& layer\_normalization, SoftmaxModule & Normalization \\
& SimpleMatmulModule, InnerProd, Transpose, Gather & Linear algebra \\
& Feedforward, PositionWiseFeedForward & Feed-forward \\
& TransformerFFNLayer, MLP\_model, MaskedLanguageModel & Networks \\
& CrossEntropyLossLabelSmoothing, KDLoss & Loss functions \\
& PositionEmbedder, GateGRUSelectionLayer & Embeddings / gating \\
\midrule
\multicolumn{3}{l}{\textbf{Triton-to-Triton -- vLLM} (118 tasks, source: vLLM inference engine)} \\
\midrule
& triton\_fused\_moe, triton\_batched\_moe, triton\_moe\_mmk & Mixture of experts \\
& triton\_flash\_prefill\_attention, triton\_unified\_attention\_* & Attention \\
& triton\_decode\_attn\_stage1/2, triton\_paged\_prefix\_prefill & Paged attention \\
& triton\_scaled\_mm, triton\_matmul\_persistent, triton\_bmm & Matrix multiply \\
& triton\_rms\_norm, triton\_layernorm\_gated, triton\_fla\_layernorm & Normalization \\
& triton\_per\_token\_group\_quant\_fp8, triton\_w8a8\_block\_* & Quantization \\
& triton\_reshape\_and\_cache\_flash, triton\_batch\_memcpy & Memory mgmt \\
& triton\_topk\_topp, triton\_gumbel\_sample, triton\_temperature & Sampling \\
& triton\_ssd\_chunk\_scan/state, triton\_selective\_scan\_update & SSM / Mamba \\
& triton\_lightning\_attn\_*, triton\_linear\_attn\_decode & Linear attention \\
& \emph{+ 78 additional kernels (see supplementary materials)} & Various \\
\midrule
\multicolumn{3}{l}{\textbf{Triton-to-Triton -- ROCmBench} (30 tasks, source: ROCmBench~\citep{geak})} \\
\midrule
& test\_add\_kernel, test\_kernel\_sub, test\_kernel\_dot, & Element-wise / \\
& test\_block\_copy, test\_load\_reduce, test\_batched\_vecmat, & reductions / \\
& test\_reverse\_range, test\_randn, test\_random\_int & data movement \\
& softmax, naive\_softmax, layernorm, rmsnorm\_fwd, rmsnorm\_bwd, & Normalization / \\
& test\_cast\_matmul, test\_chained\_matmul, test\_gemm\_no\_scf, & GEMM variants \\
& test\_iv\_dependent\_matmul, test\_triton\_sort, test\_triton\_swizzle2d & \\
& test\_flashattention\_fwd, gemm, moe\_gemm, test\_matmul\_MXFP, & Attention / \\
& test\_block\_pointer\_matmul, test\_gemm\_fusion, test\_tma\_store\_gemm, & advanced GEMM / \\
& test\_chained\_dot\_fp8, multreduce\_matmul\_dot\_kernel, & various \\
& triton\_multreduce\_matmul\_kernel & \\
\midrule
\multicolumn{3}{l}{\textbf{PyTorch-to-HIP} (24 tasks, source: GPU Mode community)} \\
\midrule
& \multicolumn{2}{l}{Same operator set as HIP-to-HIP (agent creates HIP kernel from scratch)} \\
\bottomrule
\end{tabular}}
\end{table}

\section{Example Task Directory}
\label{app:task-dir}

\begin{figure}[h]
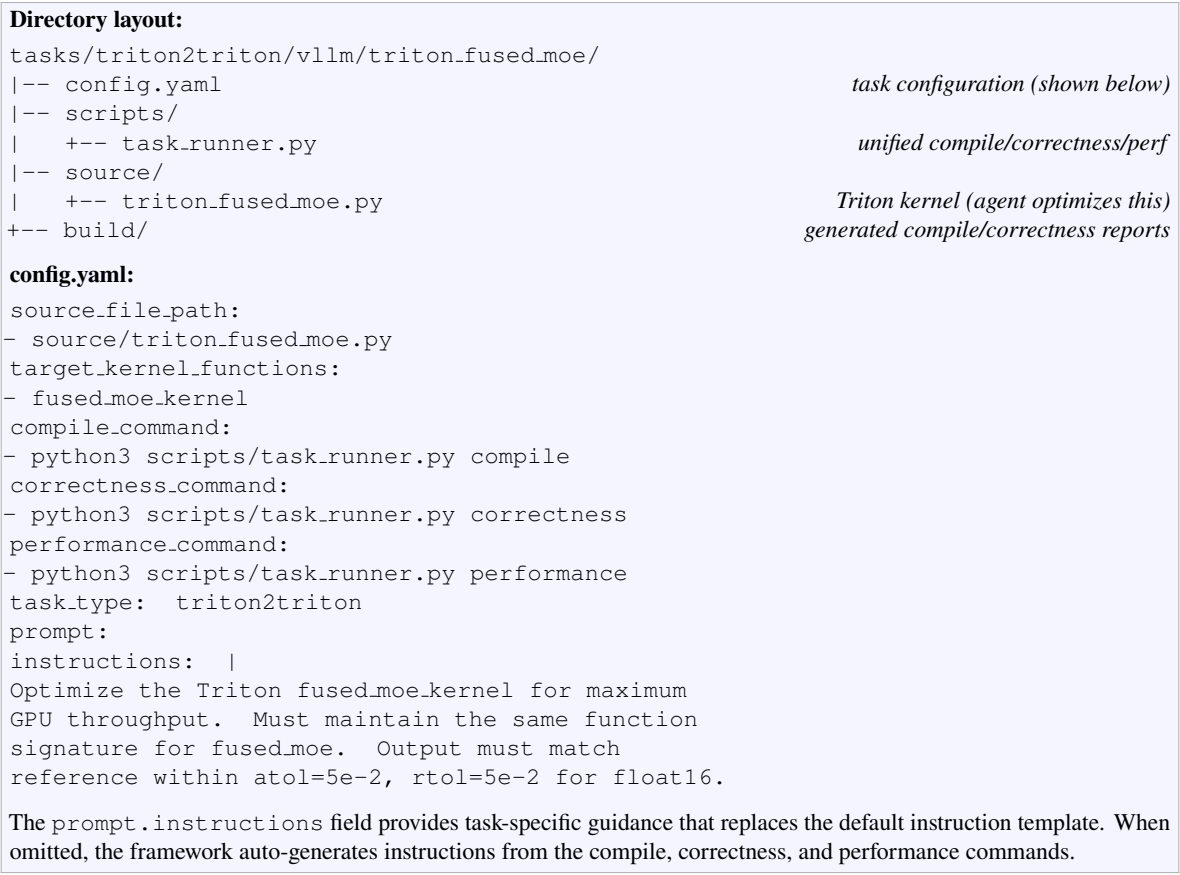

\centering
\fcolorbox{black!30}{blue!4}{%
\parbox{0.93\textwidth}{\small
\textbf{Directory layout:}\\[2pt]
\texttt{tasks/triton2triton/vllm/triton\_fused\_moe/}\\
\texttt{|-- config.yaml}\hfill\textrm{\textit{task configuration (shown below)}}\\
\texttt{|-- scripts/}\\
\texttt{|~~~+-- task\_runner.py}\hfill\textrm{\textit{unified compile/correctness/perf}}\\
\texttt{|-- source/}\\
\texttt{|~~~+-- triton\_fused\_moe.py}\hfill\textrm{\textit{Triton kernel (agent optimizes this)}}\\
\texttt{+-- build/}\hfill\textrm{\textit{generated compile/correctness reports}}\\[6pt]
\textbf{config.yaml:}\\[2pt]
\texttt{source\_file\_path:}\\
\texttt{~~- source/triton\_fused\_moe.py}\\
\texttt{target\_kernel\_functions:}\\
\texttt{~~- fused\_moe\_kernel}\\
\texttt{compile\_command:}\\
\texttt{~~- python3 scripts/task\_runner.py compile}\\
\texttt{correctness\_command:}\\
\texttt{~~- python3 scripts/task\_runner.py correctness}\\
\texttt{performance\_command:}\\
\texttt{~~- python3 scripts/task\_runner.py performance}\\
\texttt{task\_type: triton2triton}\\
\texttt{prompt:}\\
\texttt{~~instructions: |}\\
\texttt{~~~~Optimize the Triton fused\_moe\_kernel for maximum}\\
\texttt{~~~~GPU throughput. Must maintain the same function}\\
\texttt{~~~~signature for fused\_moe. Output must match}\\
\texttt{~~~~reference within atol=5e-2, rtol=5e-2 for float16.}\\[6pt]
The \texttt{prompt.instructions} field provides task-specific guidance that replaces the default instruction template. When omitted, the framework auto-generates instructions from the compile, correctness, and performance commands.
}}
\caption{Task directory layout and \texttt{config.yaml} for the \texttt{fused\_moe\_kernel} Triton-to-Triton task. The agent receives the source file to optimize, while evaluation scripts run independently after the agent session ends.}
\label{fig:task-dir}
\end{figure}

Each task is a self-contained directory with source code, evaluation scripts, and a \texttt{config.yaml} that drives prompt generation and scoring. Figure~\ref{fig:task-dir} shows the layout and configuration for the \texttt{triton\_fused\_moe} task.

\section{Agent Prompt}
\label{app:prompt}

Each agent receives a prompt assembled from seven sections. Figure~\ref{fig:prompt-example} shows the complete prompt for the \texttt{fused\_moe\_kernel} task on MI300X. The cheatsheet (section~6) is truncated; at runtime the full MI300X architecture guide (${\sim}$2\,k tokens) and Triton best-practices document (${\sim}$3\,k tokens) are appended verbatim. When a task's \texttt{config.yaml} provides a \texttt{prompt.instructions} field (as in Figure~\ref{fig:task-dir}), it replaces the default instruction template that is otherwise generated from the evaluation commands.

\begin{figure}[h]
\centering
\fcolorbox{black!30}{blue!4}{%
\parbox{0.93\textwidth}{\small
\textbf{Prompt:}\\[4pt]
\textit{\textbf{[1. Task type role]}}\\
You are a Kernel Optimization Specialist with expertise in Triton programming. Your core mission is to systematically optimize existing Triton kernels for maximum performance while ensuring strict numerical correctness and functional equivalence to the original code. You understand Triton's block-based programming model, memory tiling strategies, and how to leverage compiler hints for optimal GPU performance.\\[4pt]
\textit{\textbf{[2. Source code specification]}}\\
\textbf{File(s) to optimize:} \texttt{source/triton\_fused\_moe.py}\\
\textbf{Target kernel function(s):} \texttt{fused\_moe\_kernel}\\[4pt]
\textit{\textbf{[3. GPU architecture pre-check]}}\\
\textbf{Target GPU:} \texttt{MI300X}, architecture token: \texttt{gfx942}\\
Before running any build, test, or benchmark command, scan all build-related files for hardcoded GPU architecture strings. If any file targets an architecture other than \texttt{gfx942}, update it before proceeding.\\[4pt]
\textit{\textbf{[4. Instructions]}}\\
Optimize the Triton \texttt{fused\_moe\_kernel} for maximum GPU throughput. This is the main MoE GEMM kernel that multiplies each token by its assigned expert weight matrix using sorted token IDs and expert IDs.\\[2pt]
The kernel computes C[token] = A[token // topk] @ B[expert].T with grouped block scheduling for L2 cache reuse, and optional routing weight multiplication.\\[2pt]
Key optimization opportunities: Block size tuning (\texttt{BLOCK\_SIZE\_M}, \texttt{BLOCK\_SIZE\_N}, \texttt{BLOCK\_SIZE\_K}); \texttt{GROUP\_SIZE\_M} for L2 cache reuse; memory access patterns and prefetching; compute type selection.\\[2pt]
\textbf{Constraints:} Must maintain the same function signature for \texttt{fused\_moe}. Output must match reference within atol=5e-2, rtol=5e-2 for float16.\\[4pt]
\textit{\textbf{[5. Completion]}}\\
Save your optimized kernel code in the workspace directory. \textbf{DO NOT} write \texttt{task\_result.yaml}; the framework will automatically check compilation, validate correctness, measure performance, and generate \texttt{task\_result.yaml} with standardized metrics.\\[4pt]
\textit{\textbf{[6. Cheatsheet (truncated)]}}\\
\textbf{MI300X Architecture Guide:} CDNA3 compute topology (8 XCDs, 304 CUs, Wave64), memory hierarchy (64\,KB LDS/CU, 4\,MB L2/XCD, 256\,MB Infinity Cache, 192\,GB HBM3 at 5.3\,TB/s), MFMA instructions \ldots\\
\textbf{Triton Best Practices:} Autotuning with \texttt{@triton.autotune}, load/store masking, \texttt{tl.dot} usage, reduction strategies, AMD ROCm backend specifics \ldots\\[4pt]
\textit{\textbf{[7. Workspace directory]}}\\
Your working directory is: \texttt{/workspace/triton2triton/vllm/triton\_fused\_moe/}\\
This workspace contains all source files, build system, test/validation scripts, and profiling tools.\\[4pt]
\textit{\textbf{[8. Iteration directive (appended by the agent launcher)]}}\\
\textit{For this optimization, you must iterate up to 3 versions.}
}}
\caption{Complete prompt assembled for the \texttt{fused\_moe\_kernel} Triton-to-Triton task on MI300X. The eight sections are: (1)~task-type role, (2)~source files and target functions, (3)~GPU architecture pre-check, (4)~task-specific optimization instructions, (5)~completion directive, (6)~hardware and language cheatsheets, (7)~workspace path, and (8)~an iteration directive appended by the agent launcher when \texttt{max\_iterations} is set (default~3); this is a soft, natural-language instruction rather than a hard runtime cap on tool calls. Each task type receives a tailored role string; for example, HIP-to-HIP tasks begin with ``You are a Kernel Optimization Specialist with expertise in HIP programming.'' }
\label{fig:prompt-example}
\end{figure}

\section{Agent Configurations}
\label{app:agents}

Table~\ref{tab:agent-configs} summarizes the configuration for each evaluated agent.
All agents receive identical prompts and operate in identical sandboxed workspaces.

\begin{table}[h]
\caption{Agent configurations used in experiments.}
\label{tab:agent-configs}
\centering\small
\resizebox{\textwidth}{!}{%
\begin{tabular}{llllr}
\toprule
Agent & Interface & Models & Timeout (s) & Max Iterations \\
\midrule
Cursor Agent & CLI (cursor-agent) & Composer 2, Opus 4.6 High, Opus 4.7 High, & 3600  & 3\\
 & & GPT-5.4 High, GPT-5.3-Codex High & \\
Claude Code & CLI (claude) & Opus 4.6, Sonnet 4.6 & 3600 & 3 \\
Codex Agent & CLI (codex) &  GPT-5.3-Codex & 3600 & 3\\
\bottomrule
\end{tabular}%
}
\end{table}

All three agents operate with full shell access within the workspace and receive a prompt-level directive (\texttt{max\_iterations=3}) asking them to produce up to three successive kernel versions; this is a soft instruction, not a hard tool-call limit.
For each agent, we run experiments with multiple models to disentangle model capability from agent scaffold quality.
Each (agent, model, task type) configuration is run three times; per-task metrics are averaged across runs before computing aggregate statistics.

\paragraph{Model versions and evaluation dates.}
Commercial LLM providers update the weights behind stable model identifiers without versioning, so identifier strings alone are insufficient to pin down the model that produced a given result.
To allow temporal model drift to be accounted for, we report both the exact identifier string passed to each agent platform and the calendar window in which each configuration was executed (Table~\ref{tab:model-versions}).
All experiments were conducted between April 16 and April 24, 2026 on AMD Instinct MI300X.
Where an identifier carries a Cursor-internal compute tier (e.g., \texttt{-high}), this is the platform-facing name and is reproducible only inside the Cursor Agent scaffold; \texttt{composer-2} is likewise a Cursor-internal model.
Unless otherwise noted, all other sampling parameters (temperature, top-$p$, retry policy) were left at each platform's defaults.

\begin{table}[h]
\caption{Model identifiers and evaluation windows for each agent configuration. Identifiers are the exact strings passed to the agent platform. Cursor-internal tier suffixes (\texttt{-high}) and Cursor-exclusive models (\texttt{composer-2}) are reproducible only inside the Cursor Agent scaffold.}
\label{tab:model-versions}
\centering\small
\begin{tabular}{lll}
\toprule
Agent & Model identifier & Evaluation window (2026) \\
\midrule
Cursor Agent & \texttt{claude-4.6-opus-high}     & Apr 19--20 \\
Cursor Agent & \texttt{claude-opus-4-7-high}     & Apr 21--22 \\
Cursor Agent & \texttt{gpt-5.4-high}             & Apr 22--24 \\
Cursor Agent & \texttt{gpt-5.3-codex-high}       & Apr 17--20 \\
Cursor Agent & \texttt{composer-2}               & Apr 15--16 \\
\midrule
Claude Code  & \texttt{claude-opus-4-6}          & Apr 16--17 \\
Claude Code  & \texttt{claude-sonnet-4-6}        & Apr 16--17 \\
\midrule
Codex Agent  & \texttt{gpt-5.3-codex}            & Apr 20--24 \\
\bottomrule
\end{tabular}
\end{table}

\paragraph{Unseen configuration generation.}
The unseen configurations used in \S\ref{sec:generalization} were produced in a separate, one-shot pipeline using Cursor Agent with \texttt{claude-4.6-opus-high}; this configuration is distinct from the agents evaluated in the main results and is not part of any reported speedup measurement.
The generated configuration definitions are not redistributed, to prevent contamination of future evaluations, while the generation script is released as part of the benchmark repository so that researchers can reproduce the protocol, inspect its biases, and extend it to new task categories.

\section{Extensibility Guide}
\label{app:extensibility}

AgentKernelArena is designed for easy extension along three axes.

\paragraph{Adding a new agent.}
A new agent requires two files under \texttt{agents/<name>/}:
\begin{itemize}[leftmargin=*]
  \item \texttt{launch\_agent.py}: a Python module that registers a launcher function via the \texttt{@register\_agent} decorator. The launcher receives three arguments (the global config, the task config path, and the workspace path) and is responsible for invoking the agent (via subprocess, API call, etc.) within the workspace.
  \item \texttt{agent\_config.yaml}: agent-specific settings such as model name, timeout, and any agent-specific parameters.
\end{itemize}
The agent is then selectable via the global \texttt{config.yaml} by setting \texttt{agent.template} to the registered name.
No changes to the evaluation pipeline are required: the centralized evaluator handles all scoring independently.

\paragraph{Adding a new task.}
A new task is a directory under \texttt{tasks/<category>/<name>/} containing:
\begin{itemize}[leftmargin=*]
  \item \texttt{config.yaml}: specifies source files, target kernel functions, compile/correctness/performance commands, task type, and optional prompt overrides.
  \item Source files: the kernel(s) to optimize and any reference implementations.
  \item Evaluation scripts: task-specific compilation, correctness checking, and performance measurement scripts.
\end{itemize}
The framework automatically discovers tasks via filesystem glob and includes them in runs based on the \texttt{tasks} list in the global config.

\paragraph{Adding a new GPU architecture.}
Hardware support is configured in \texttt{src/prompts/cheatsheet/default\_cheatsheet.yaml}, which maps GPU model names to architecture tokens (e.g., MI300X $\to$ \texttt{gfx942}) and cheatsheet files.
Supporting a new GPU requires: (i)~adding an architecture entry to this YAML, (ii)~writing an architecture guide markdown file, and (iii)~optionally adding language-specific best practices.
The \texttt{target\_gpu\_model} field in the global config selects the active architecture.
The framework already includes entries for both MI300X and MI355X.

\section{Speedup Distribution Details}
\label{app:speedup-dist}

Table~\ref{tab:dist} reports the cross-task speedup distribution (std, P25, P75, P90) for each configuration, complementing the mean and geometric mean in the main tables.
Note that this cross-task std captures variance \emph{across tasks} within a single configuration (some kernels offer more optimization headroom than others), which is distinct from the run-to-run $\sigma_r$ reported in the main tables.
The high Triton-to-Triton std values (6--9$\times$ despite mean speedups near 2$\times$) reflect a small number of tasks with exceptionally high speedup headroom that inflate the cross-task variance.

\begin{table}[h]
\caption{Cross-task speedup distribution per configuration (computed over per-task run-averaged speedups).}
\label{tab:dist}
\centering\small
\resizebox{0.9\textwidth}{!}{%
\begin{tabular}{lllrrrr}
\toprule
Category & Agent & Model & Task Std & P25 & P75 & P90 \\
\midrule
\multirow{8}{*}{HIP-to-HIP}
 & Claude Code  & Opus 4.6       &  8.70$\times$ & 1.18$\times$ &  7.01$\times$ & 19.96$\times$ \\
 & Claude Code  & Sonnet 4.6     &  6.60$\times$ & 1.04$\times$ &  6.03$\times$ & 17.55$\times$ \\
 & Cursor Agent & Opus 4.7 High  &  7.83$\times$ & 1.22$\times$ &  5.73$\times$ & 19.11$\times$ \\
 & Cursor Agent & Opus 4.6 High  &  6.89$\times$ & 1.07$\times$ &  3.12$\times$ & 15.35$\times$ \\
 & Cursor Agent & GPT-5.4 High   &  6.51$\times$ & 1.04$\times$ &  6.54$\times$ & 15.14$\times$ \\
 & Cursor Agent & GPT-5.3-Codex High &  4.55$\times$ & 1.08$\times$ &  2.66$\times$ & 12.94$\times$ \\
 & Cursor Agent & Composer 2     &  0.94$\times$ & 1.02$\times$ &  1.53$\times$ &  2.17$\times$ \\
 & Codex Agent  & GPT-5.3-Codex  &  4.08$\times$ & 1.09$\times$ &  3.86$\times$ & 10.88$\times$ \\
\midrule
\multirow{8}{*}{\shortstack[l]{Triton-to-\\Triton}}
 & Claude Code  & Opus 4.6       &  8.26$\times$ & 1.01$\times$ & 1.34$\times$ & 1.88$\times$ \\
 & Claude Code  & Sonnet 4.6     &  7.95$\times$ & 1.00$\times$ & 1.36$\times$ & 1.89$\times$ \\
 & Cursor Agent & Opus 4.7 High  &  9.29$\times$ & 1.01$\times$ & 1.45$\times$ & 1.95$\times$ \\
 & Cursor Agent & Opus 4.6 High  &  6.58$\times$ & 1.00$\times$ & 1.23$\times$ & 1.41$\times$ \\
 & Cursor Agent & GPT-5.4 High   &  7.81$\times$ & 0.98$\times$ & 1.04$\times$ & 1.22$\times$ \\
 & Cursor Agent & GPT-5.3-Codex High &  7.64$\times$ & 0.96$\times$ & 1.03$\times$ & 1.18$\times$ \\
 & Cursor Agent & Composer 2     &  6.86$\times$ & 0.87$\times$ & 1.05$\times$ & 1.47$\times$ \\
 & Codex Agent  & GPT-5.3-Codex  &  7.52$\times$ & 0.98$\times$ & 1.03$\times$ & 1.12$\times$ \\
\midrule
\multirow{8}{*}{\shortstack[l]{PyTorch-\\to-HIP}}
 & Claude Code  & Opus 4.6       &  6.99$\times$ & 2.99$\times$ & 7.17$\times$ & 14.77$\times$ \\
 & Claude Code  & Sonnet 4.6     &  5.31$\times$ & 2.44$\times$ & 6.47$\times$ & 10.55$\times$ \\
 & Cursor Agent & Opus 4.7 High  &  5.97$\times$ & 2.90$\times$ & 8.89$\times$ & 15.53$\times$ \\
 & Cursor Agent & Opus 4.6 High  &  6.91$\times$ & 2.38$\times$ & 8.55$\times$ & 15.73$\times$ \\
 & Cursor Agent & GPT-5.4 High   &  3.81$\times$ & 1.00$\times$ & 6.91$\times$ &  9.00$\times$ \\
 & Cursor Agent & GPT-5.3-Codex High &  3.11$\times$ & 1.39$\times$ & 4.18$\times$ &  8.72$\times$ \\
 & Cursor Agent & Composer 2     &  3.90$\times$ & 1.80$\times$ & 4.34$\times$ &  6.73$\times$ \\
 & Codex Agent  & GPT-5.3-Codex  &  4.26$\times$ & 2.22$\times$ & 7.57$\times$ & 11.93$\times$ \\
\bottomrule
\end{tabular}%
}
\end{table}

\section{Agent Behavior Analysis}
\label{app:behavior}

We analyze orchestration logs from representative runs across all task categories to quantify agent behavior.

\paragraph{Iteration and completion.}
On HIP-to-HIP (24 tasks) and PyTorch-to-HIP (24 tasks), all agents complete every task within the timeout.
On Triton-to-Triton (148 tasks), all evaluated agents consistently complete the full task set.

\paragraph{Compilation and correctness failures.}
Failure patterns differ markedly across task types.
PyTorch-to-HIP shows the highest compilation churn: Claude Code encounters compilation failures on approximately 35 occasions across 24 tasks (multiple retries per task), while Composer~2 surfaces zero compile-failure strings, suggesting it validates code more carefully before attempting compilation.
Triton tasks exhibit few compilation failures but correctness failures tied to autotuning: misconfigured \texttt{@triton.autotune} parameters (e.g., missing \texttt{reset\_to\_zero} for atomic kernels) cause silent numerical errors that agents must diagnose and revert.
On HIP-to-HIP, 1--2 tasks per run pass the agent's internal checks but fail centralized correctness, indicating occasional tolerance violations.

\paragraph{Optimization strategies by task type.}
HIP-to-HIP agents focus on memory access optimization (coalescing, vectorization), compute fusion, and AMD CDNA3-specific features (\texttt{\_\_launch\_bounds\_\_}, shared memory management).
Triton-to-Triton agents primarily manipulate autotuning configurations: \texttt{BLOCK\_SIZE}, \texttt{num\_warps}, \texttt{num\_stages}, and AMD-specific \texttt{waves\_per\_eu} and \texttt{matrix\_instr\_nonkdim}.
PyTorch-to-HIP agents must solve the additional challenge of mapping PyTorch module semantics to HIP kernel launches, including thread-block decomposition, memory allocation (\texttt{hipMalloc}), and Python bindings via \texttt{torch.utils.cpp\_extension}.

\section{Failure Case Analysis}
\label{app:failures}

We categorize the failure modes observed across all evaluation runs and identify which agent/model configurations are most susceptible to each type.

\paragraph{PyTorch-to-HIP compilation failures.}
Compilation failures occur almost exclusively in PyTorch-to-HIP tasks, where the agent must generate a complete HIP kernel with Python bindings from scratch. The dominant failure mode is a missing or malformed \texttt{PYBIND11\_MODULE} entry point, causing the compiled shared library to lack the required \texttt{PyInit\_*} symbol:

\begin{center}
\fcolorbox{red!30}{red!3}{\parbox{0.88\textwidth}{\small\ttfamily
ImportError: dynamic module does not define module export function (PyInit\_hip\_11178\_TanH)
}}
\end{center}

\noindent This failure is strongly model-dependent. Cursor Agent with GPT-5.4 High has the lowest PyTorch-to-HIP compilation rate (${\sim}$70\%, failing on 7 of 24 tasks per run, including \texttt{SiLU}, \texttt{TanH}, \texttt{Sigmoid}, \texttt{Gather}, \texttt{Transpose}, \texttt{MultiHeadAttention}, and \texttt{layer\_normalization}). By contrast, Claude Code (Opus 4.6, Sonnet 4.6), Cursor Agent with Opus 4.6/4.7 High, Cursor Agent with Composer 2, and Codex Agent all achieve 95--100\% compilation rates on the same tasks. This suggests that the ability to correctly wire \texttt{torch.utils.cpp\_extension.load()} bindings varies substantially across underlying LLMs.

\paragraph{HIP-to-HIP correctness failures.}
HIP-to-HIP tasks have near-perfect correctness across all configurations: all agents achieve $\geq$91.7\% correctness (typically 23--24 out of 24 tasks). The rare failures (1--2 per run at most) involve tolerance violations where the optimized kernel produces numerically acceptable but not bitwise-identical results. These occur sporadically across models without a clear pattern, confirming that HIP kernel optimization, where the agent modifies existing working code, is a more forgiving task than generating code from scratch.

\paragraph{Triton-to-Triton correctness failures.}
Triton correctness is high overall (96--100\%) but shows a model-specific pattern. Cursor Agent with Composer~2 has the most Triton correctness failures (3--5 per run out of 148 tasks, ${\sim}$97\%), while Cursor Agent with Opus 4.6/4.7 High and Claude Code with Opus 4.6 achieve perfect 100\% correctness across all three runs. Failures fall into two categories.

The first involves type mismatches in conditional branches, where the agent introduces inconsistent tensor types across control flow paths:

\begin{center}
\fcolorbox{orange!40}{orange!3}{\parbox{0.88\textwidth}{\small\ttfamily
CompilationError: Mismatched type for a between then block (<['64'], int1>) and else block (<['64'], int8>)
}}
\end{center}

\noindent Although the kernel compiles for standard input configurations, it fails at Triton's JIT compilation stage for specific parameter combinations (e.g., boolean dtypes with zero-padding), causing the centralized evaluator to record a correctness failure.

The second category involves numerical precision issues with specialized data types. On the \texttt{moe\_gemm} task, the agent's optimization passes all standard float16 tests but fails on FP8 test cases, where tighter numerical tolerances expose rounding differences introduced by the optimization.

\paragraph{Unseen-configuration generalization failures.}
The unseen-input evaluation reveals a distinct and important failure mode: agents that achieve perfect correctness on original test shapes fail on unseen shapes due to hardcoded assumptions. \textbf{PyTorch-to-HIP is the most affected category} (54--92\% unseen-input retention depending on model).

\textit{PyTorch-to-HIP: the most affected category.}
Because agents generate HIP kernels from scratch, they frequently hardcode buffer sizes, thread-block dimensions, and loop bounds derived from the original test shapes. Two representative examples:

\begin{center}
\fcolorbox{blue!30}{blue!3}{\parbox{0.88\textwidth}{\small
\textbf{torch2hip / Feedforward} (Cursor / Opus 4.7 High, original speedup: 9.6$\times$, unseen: FAIL):\\[1pt]
\texttt{[Error] Feedforward raises an exception due to total rows (64) exceeds MAX\_ROWS (32).}
}}
\end{center}
\vspace{4pt}
\begin{center}
\fcolorbox{blue!30}{blue!3}{\parbox{0.88\textwidth}{\small
\textbf{hip2hip / KDLoss} (Claude Code / Opus 4.6, original speedup: 28.9$\times$, unseen: FAIL):\\[1pt]
\texttt{[Error] KDLoss raises an exception due to Channel dim C=37 exceeds MAX\_C=32.}
}}
\end{center}

\noindent In both cases, the agent hardcoded a constant (\texttt{MAX\_ROWS=32}, \texttt{MAX\_C=32}) in shared memory allocations based on the original test shapes. Unseen shapes with larger dimensions exceeded these constants, causing runtime failures despite 100\% correctness on original shapes.

Cursor Agent / GPT-5.4 High is the most affected configuration, with 8--11 of 24 tasks regressing on unseen shapes (54--67\% retention). Cursor Agent / Opus~4.6 High performs best (92\% retention), followed by Codex Agent / GPT-5.3-Codex (88--92\%). This ordering largely mirrors the original-run compilation rates, suggesting that models prone to binding errors are also more likely to hardcode shape-specific constants.

\textit{Triton-to-Triton unseen-input failures.}
Triton tasks show fewer regressions (90--100\% retention) but with distinct mechanisms:

\begin{center}
\fcolorbox{blue!30}{blue!3}{\parbox{0.88\textwidth}{\small
\textbf{triton2triton / count\_expert\_tokens} (Cursor / Composer 2):\\[1pt]
\texttt{Shape element 0 must be a power of 2}
}}
\end{center}

\noindent The agent used \texttt{tl.histogram} with a \texttt{NUM\_BINS} parameter that happened to be a power of 2 for all original shapes, but an unseen configuration introduced a non-power-of-2 expert count, violating Triton's constraint.
On the \texttt{fla\_fused\_recurrent} task, Claude Code / Opus~4.6's optimization passes original shapes but produces \texttt{max diff = 4.70} on an unseen shape, indicating that the tiling strategy accumulates numerical error at different sequence lengths.

\textit{HIP-to-HIP unseen-input failures.}
HIP-to-HIP tasks have the highest unseen-input retention (91--100\%), since agents modify existing working code rather than generating from scratch, limiting opportunities to introduce shape-specific assumptions.

\paragraph{Implications for benchmark design.}
These failure modes validate three design choices: (1)~centralized evaluation catches failures the agent's own checks may miss, since agents typically test only a subset of input configurations; (2)~multi-shape, multi-dtype test cases expose correctness issues that single-configuration testing would not detect; and (3)~unseen-configuration generalization testing reveals hardcoded assumptions that inflate reported speedups.

\section{Token Usage Breakdown}
\label{app:tokens}

\begin{table}[h]
  \caption{Average LLM output tokens per task (thousands), averaged over 3 runs. Per-task counts are obtained by dividing per-run totals by the number of tasks in each category (24 for HIP-to-HIP and PyTorch-to-HIP, 148 for Triton-to-Triton).}
  \label{tab:tokens}
  \centering\small
  \begin{tabular}{llrrr}
    \toprule
    Agent & Model & HIP-to-HIP & Triton-to-Triton & PyTorch-to-HIP \\
    \midrule
    Claude Code  & Opus 4.6           & 59.0 & 48.8 & 38.9 \\
    Claude Code  & Sonnet 4.6         & 86.4 & 58.9 & 40.5 \\
    \midrule
    Cursor Agent & Opus 4.7 High      & 25.0 & 17.7 & 20.8 \\
    Cursor Agent & Opus 4.6 High      & 19.7 & 18.9 & 13.7 \\
    Cursor Agent & GPT-5.4 High       & 10.3 & 16.8 &  8.8 \\
    Cursor Agent & GPT-5.3-Codex High & 15.4 & 12.4 & 12.9 \\
    Cursor Agent & Composer 2         & 12.9 &  8.3 & 10.8 \\
    \midrule
    Codex Agent  & GPT-5.3-Codex      & 16.8 & 13.7 & 12.7 \\
    \bottomrule
  \end{tabular}
\end{table}

\section{Example Agent Interaction}
\label{app:transcript}

Below is an abridged transcript of a Claude Code session optimizing a Triton fused MoE kernel (\texttt{triton\_fused\_moe}), illustrating the iterative compile--test--profile loop.

\begin{small}
\begin{verbatim}
[Turn 1] Agent reads source/triton_fused_moe.py
         Agent identifies fused_moe_kernel as target
         Agent runs: python3 scripts/task_runner.py compile
         > Compilation successful

[Turn 2] Agent runs: python3 scripts/task_runner.py correctness
         > All 5 test cases passed
         Agent runs: python3 scripts/task_runner.py performance
         > Test case 1: 0.342 ms, Test case 2: 1.205 ms, ...

[Turn 3] Agent modifies BLOCK_SIZE_M=128, BLOCK_SIZE_K=64
         Agent adds tl.dot accumulator optimization
         Agent runs: python3 scripts/task_runner.py compile
         > Compilation successful

[Turn 4] Agent runs: python3 scripts/task_runner.py correctness
         > Test case 3 FAILED (atol exceeded)
         Agent reverts BLOCK_SIZE_K change, adjusts masking

[Turn 5] Agent runs correctness again
         > All 5 test cases passed
         Agent runs performance
         > Test case 1: 0.298 ms (-12.9%), Test case 2: 0.987 ms (-18.1%), ...

[Turn 6] Agent tries shared memory prefetching
         Compilation fails: "shared memory exceeds LDS limit"
         Agent reduces tile size, recompiles successfully
         Correctness: passed, Performance: further 5% improvement

[Framework] Agent session ends (timeout or completion)
            Centralized evaluator re-runs compile/correctness/performance
            Final speedup: 1.23x average across 5 test shapes
            Score: 20 + 100 + 123 = 243
\end{verbatim}
\end{small}

\section{Limitations}
\label{app:limitations}

The current study targets a single GPU architecture (AMD MI300X) and evaluates three commercial agents over three runs per configuration, limited by API cost. Model availability varies across platforms, so most model comparisons use Cursor Agent. Open-weight models were explored with single-iteration calls but consistently failed at compilation due to the large multi-file contexts involved; designing iterative feedback loops for them was outside the scope of this benchmark. Specialized kernel optimization systems (e.g., GEAK, AutoTriton) were also excluded, as their task-specific architectures differ from general-purpose coding agents and a fair comparison would not be possible in this study, though the framework readily supports their integration. The task set is drawn primarily from vLLM and GPU Mode, and we are actively expanding it with kernels from repositories such as AITER.


\end{document}